\documentclass[letterpaper, 10 pt, journal, twoside]{ieeetran}
\IEEEoverridecommandlockouts                              % This command is only needed if 
\usepackage{amsmath,mathtools,amsfonts}
\usepackage{graphicx}
\usepackage{multirow}
\usepackage{booktabs}
\usepackage{algorithm}
\usepackage{algpseudocode}
\usepackage{physics}
\usepackage{subfig}
\usepackage{wrapfig}
\usepackage{hyperref}
\usepackage{xcolor}

\usepackage{cite}
\definecolor{green}{RGB}{3,112,15}

\DeclarePairedDelimiter\floor{\lfloor}{\rfloor}

\DeclareMathOperator*{\argmax}{argmax}
\DeclareMathOperator*{\argmin}{argmin}

\title{A Lifelong Learning Approach to \\Mobile Robot Navigation}

\author{Bo Liu$^{1*}$, Xuesu Xiao$^{1*}$, and Peter Stone$^{1, 3}$
\thanks{Manuscript received: October, 15, 2020; Revised December, 20, 2020;
Accepted January, 24, 2021.}%Use only for final RAL version
\thanks{This paper was recommended for publication by Editor Dan Popa upon
evaluation of the Associate Editor and Reviewers' comments. 
}%Use only for final RAL version
\thanks{$^{*}$Equally contributing authors}
\thanks{$^{1}$Bo Liu, Xuesu Xiao, and Peter Stone are with Department of Computer Science, The University of Texas at Austin, Austin, TX 78712 {\tt\small \{xiao, bliu, pstone\}@cs.utexas.edu}}
\thanks{$^{3}$Peter Stone is with Sony AI}%
\thanks{Digital Object Identifier (DOI): see top of this page.}
}
\begin{document}
\maketitle
% \thispagestyle{empty}
% \pagestyle{empty}

%%%%%%%%%%%%%%%%%%%%%%%%%%%%%%%%%%%%%%%%%%%%%%%%%%%%%%%%%%%%%%%%%%%%%%%%%%%%%%%%
%===============================================================================

\begin{abstract}
% objective with no jargon
This paper presents a self-improving lifelong learning framework for a mobile robot navigating in different environments.
% how is it done today 
Classical static navigation methods require environment-specific in-situ system adjustment, e.g. from human experts, or may repeat their mistakes regardless of how many times they have navigated in the same environment. Having the potential to improve with experience, learning-based navigation is highly dependent on access to training resources, e.g. sufficient memory and fast computation, and is prone to forgetting previously learned capability, especially when facing different environments.
% what's new
In this work, we propose Lifelong Learning for Navigation (LLfN) which (1) improves a mobile robot's navigation behavior purely based on its own experience, and (2) retains the robot's capability to navigate in previous environments after learning in new ones. LLfN is implemented and tested entirely onboard a physical robot with a limited memory and computation budget. 
% who cares
% To the best of the authors' knowledge, this is the first self-supervised lifelong learning framework for mobile robot locomotion.
\end{abstract}

\section{Introduction}
\IEEEPARstart{C}{lassical} mobile robots are designed to be adaptive to different navigation environments by in-situ adjustment of the underlying navigation system, such as by sensor calibration \cite{xiao2017uav} or by parameter tuning \cite{xiao2020appld}. However, without adjustment from expert knowledge, the untuned system may repeat the same mistakes (e.g. stuck in the same bottleneck) even though it has navigated in the same environment multiple times. 

Recent success in using machine learning for mobile robot navigation indicates the potential of improving navigation performance from a robot's past experience in the same environment \cite{kahn2018self}. When facing different navigation environments, however, learning methods cannot generalize well to unseen scenarios: They must re-learn to navigate in the new environments. More importantly, the learned system is prone to \emph{catastrophic forgetting}, which causes the robot to forget what was learned in previous environments \cite{french1999catastrophic}. 

This paper introduces a Lifelong Learning for Navigation (LLfN) framework that addresses the aforementioned challenges: Instead of learning from scratch, the navigation policy is initialized through a classical navigation algorithm, whose navigation performance does not improve with increasing experience. The robot is able to identify its suboptimal actions and learn from them. The navigation performance then improves in a self-supervised manner. When facing different navigation environments, the navigation policy is able to learn to adapt to new environments, while not forgetting how to navigate in previous ones. LLfN is implemented entirely onboard a physical robot with limited memory and computation, and demonstrated to allow the robot to navigate in three different environments (Figure \ref{fig::three_tasks}). The main contributions of the paper are: 
% \peter{Yeah, especially with how much you refer to “tasks” in the intro, it needs to be clarified.}

\begin{figure}[t]
\centering
\includegraphics[width=\columnwidth]{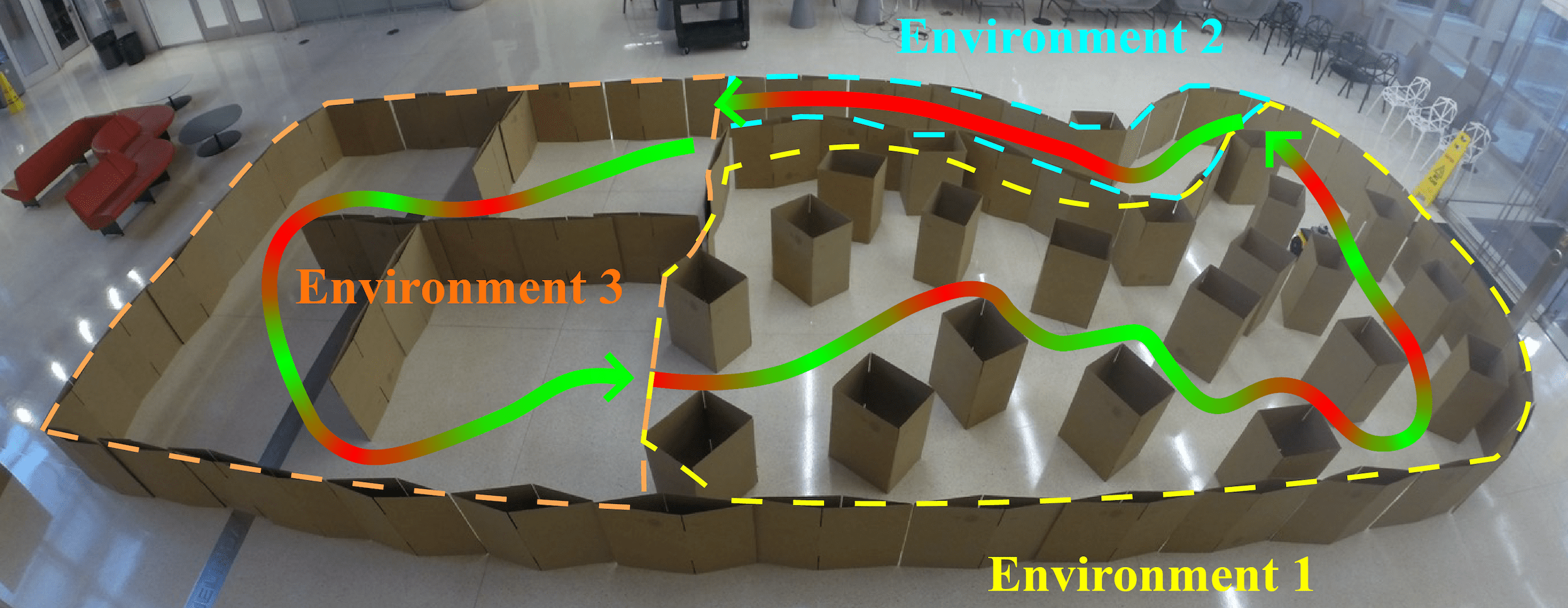}
\caption{Three navigation environments: An initial navigation policy navigates well most of the time ({\color{green} green}), but occasionally behaves suboptimally ({\color{red}red}, e.g. moving extremely slowly or getting stuck). Lifelong Learning for Navigation learns a complementary policy deployed in conjunction with the initial policy, which gradually eliminates the suboptimal behaviors in the current environment while not diminishing performance in previous environments. During deployment, the learned policy is mostly used in the red segments. }
\label{fig::three_tasks}
\end{figure}

\begin{itemize}
    \item A self-improvement strategy that complements an initial static planner to dynamically increase navigation performance with more experience, deployed in conjunction with the initial planner to minimize learning overhead;
    \item A lifelong learning scheme that allows a robot to navigate in new environments while not forgetting previous ones; and
    \item An implementation of the Lifelong Learning for Navigation framework entirely onboard a physical robot platform with limited memory and computation.  
\end{itemize}

%===============================================================================
\section{Related Work}
\label{sec::related_work}
This section first reviews how classical and learning-based methods improve navigation performance and adapt to different navigation environments, and then briefly discusses recent successes in the continual learning community.

% \subsection{Improving and Adapting Navigation}
% Classical navigation often relies on parameter tuning to improve performance and adapt to different navigation environments, while existing learning methods only learn navigation in a single-environment setting. 
% \peter{It might actually make sense to put the notation section before related work so that you can formally define “task” (as opposed to just informally in the intro) prior to the related work section (since you use the concept again a bunch).  If the explanation in the intro is clear enough, it can probably wait until Sec. 3.}

\paragraph{Classical Navigation} Classical navigation systems \cite{quinlan1993elastic, fox1997dynamic, kavraki1996probabilistic, kuffner2000rrt} are designed to be applicable to a wide variety of environments, but they are usually static, operating under a fixed set of pre-specified hyper-parameters and therefore lacking the ability to improve with experience and adapt to a specific environment. 
Parameter Tuning is the current practice to address the aforementioned problems \cite{zheng2017ros}, which requires human experts' intuition, experience, and trial-and-error. 
%
% An automated parameter tuning approach has been proposed by \cite{binch2020context}, in which two sets of navigation parameters are tuned through generic algorithms to perform straight-line and U-turn navigation. 
%
To reduce the reliance on expert knowledge, Xiao {\em et al.} \cite{xiao2020appld} proposed to improve navigation for a given environment (defined as ``context'') by tuning parameters through Behavior Cloning (BC) from teleoperated demonstration.
%
% In classical navigation, improvement and adaptation are often simultaneously addressed by hyper-parameter tuning in the target environment. 
In most cases, tuning requires human knowledge, either in the form of direct tuning or of navigation demonstrations. Furthermore, once tuned, the static navigation system lacks the ability to further improve with more experience or adapt to new environments. 
In contrast, the proposed Lifelong Learning for Navigation does not require human knowledge and can dynamically improve with more navigation experience when facing new environments. 

\paragraph{Learning-based Navigation} Data-driven machine learning techniques have also been widely applied to navigation problems~\cite{pfeiffer2018reinforced, tai2017virtual, zeng2019navigation, wang2018learning, xie2018learning, chiang2019learning, gao2017intention, zhu2017target, wang2019dual}. As for physical robot navigation, learning approaches typically either imitate an expert~\cite{pfeiffer2017perception,silver2010applied} or learn from trial-and-error using reinforcement learning~\cite{kahn2018self,han2018sensor}.
While these learning methods enable improvement in a specific environment with increased navigation experience, if the agent were to be placed in multiple environments in a sequential fashion, which is common in real-world navigation, learning methods may not generalize well and can easily forget the past knowledge. By contrast, LLfN explicitly considers how to prevent forgetting and increase generalization.
From a lifelong learning perspective, Wyeth and Milford \cite{wyeth2011towards} studied continual mapping for navigation and Wang {\em et al.} \cite{wang2019reinforced} improved generalization in vision-language navigation. Both methods focus on the problem of \emph{where} to navigate instead of \emph{how} to navigate. To the best of the authors' knowledge, no existing work has tackled lifelong/continual learning of navigation behaviors across different navigation environments. We conjecture that unconstrained offboard computation resources, e.g., memory, power, and time, allow learning from an extensive body of training data pre-collected in different environments. However, for onboard resource-constrained robot platforms without pre-collected supervised data, learning how to navigate in new environments while not forgetting previous ones requires further research, and is the focus of this work. 
% Machine learning techniques have also been applied to mobile robot navigation. These data-driven approaches have the potential of improvement with more experience (data). 
% %
% \cite{pfeiffer2017perception} used imitation learning to learn from demonstration provided by a classical planner, Dynamic Window Approach (DWA) \cite{fox1997dynamic}, performing navigation in a simple environment. The learned navigation improves with more demonstrated data. 
% %
% Another approach to improve with experience is Reinforcement Learning (RL), which learns from trial-and-error. \cite{kahn2018self} used RL to learn to navigate a closed-loop hallway in a collision-free manner.  The number of collisions can be reduced with more learning experience.
% %
% These works conducted training offboard the robot, which requires GPU and hours of offboard computation. 
% %
% To the best of our knowledge, no existing work has tackled lifelong/continual learning of navigation behaviors across different navigation environments. We conjecture that unconstrained offboard computation resources, e.g., memory, power, and time, allow learning from an extensive body of training data collected in different environments. However, for onboard resource-constrained robot platforms, learning to navigate in new environments while not forgetting previous ones requires further research, which is the focus of this work. 

\paragraph{Lifelong/Continual Supervised Learning}
Lifelong or continual learning studies the problem of learning in an ongoing fashion. One of the earliest attempts at lifelong learning originates from the robotics community~\cite{thrun1995lifelong}. Ring \cite{ring1994continual} provides the earliest introduction to continual learning in reinforcement learning problems. Recently, much progress has been made for continual learning with neural networks. There are mainly three categories of approaches: 1) use regularization to prevent the learned weights from deviating too much from the old weights~\cite{kirkpatrick2017overcoming,zenke2017continual}; 2) train a generative model to recover old data for joint optimization~\cite{shin2017continual}; and 3) adopt a dynamic network architecture for learning more tasks~\cite{rusu2016progressive,yoon2017lifelong}. Among the above approaches, the first approach applies to a fixed capacity network, which is often much more computationally efficient than training a generative model or adopting a dynamic network architecture. This computational efficiency is essential for learning onboard resource-constrained mobile robot platforms. Specifically, when a few past data points can be saved, Gradient Episodic Memory (GEM)~\cite{lopez2017gradient} can be very efficient and powerful. All these methods demonstrate success on continual image classification problems, but there remains much room for studying continual learning in other applications like in robotics,
%\commentp{I don't think we need to say robotics is more "realistic" than image classification.  some people may take issue with that, and it's not important for our message.}
especially when supervised labels from human experts are not available a priori. 
It is worth to note lifelong learning's resemblance to transfer learning~\cite{taylor2009transfer}. However, in addition to transfer learning's focus on forward transfer (how previous knowledge can help learning the current task), LLfN also considers backward transfer (how learning the current task can maintain or improve performance of old ones).
\section{Background}
\label{sec::background}
Lifelong Learning for Navigation (LLfN) aims to address a novel variant of the standard navigation problem in which the agent learns to improve navigation performance online with increasing experience, or across environments, under a limited memory budget.
We think of this problem variant as the ``lifelong navigation" problem.\footnote{We define ``lifelong navigation" in the sense of learning navigation with increasing experience, across environments, rather than in the literal sense, i.e. navigation over extended periods of time.}
% \footnote{We define ``lifelong navigation" in the sense of learning navigation with increasing experience, across environments, rather than in the literal sense, i.e. navigation over extended periods of time.}
In this section, we first present the problem setup of lifelong navigation and the notation we use in this work. We then leverage a lifelong learning algorithm previously used in continual image classification for the LLfN framework.

\subsection{Problem Setup and Notation}
\label{subsec::problem}
The high-level objective of lifelong navigation can be summarized as learning to navigate in a sequence of $m$ environments $\{\mathcal{E}_i\}_{i=1}^m$.\footnote{Informally, an environment is a contiguous space consists of similar distribution of obstacles where the optimal navigation behavior does not vary much. We intentionally design the simulated and physical environments to be very different to magnify the effect with and without LLfN.}
% \footnote{Informally, we consider an environment to be a contiguous space in which a fixed navigation policy without any adaptation (e.g. parameter-tuning or re-learning) will achieve similar performance. It is generally characterized by the width of its passageways and the density of its obstacles.}
In each of those environments, the robot aims at navigating from one fixed start point to another fixed goal point. We assume a fixed global planner (e.g. Dijkstra's algorithm \cite{dijkstra1959note}, A* \cite{Hart1968} or D* \cite{ferguson2006using}) generates a path connecting start and goal, and while navigating, the robot needs to produce motion commands which follow this global path, observe its kinodynamic constraints, and avoid obstacles. 
Whenever the agent advances to $\mathcal{E}_k$, it no longer has access to $\{\mathcal{E}\}_{i=1}^{k-1}$. Within the environment $\mathcal{E}_k$, the agent, at each time step $t$, computes a motion command $a_t \in \mathcal{A} \sim \pi_\theta(s_t)$, where $s_t \in \mathcal{S}$ is the agent's state, $\pi_\theta$ is a policy parameterized by $\theta$.
After executing $a_t$, the agent advances to $s_{t+1}$ and the process continues. During the learning phase, the agent will have a limited onboard memory size $n$, i.e. maximally $n$ pairs of $(s_t, a_t)$ can be stored at any moment. Once the agent has seen all $m$ environments, its navigation performance will be evaluated on the same $m$ environments.

\subsection{Gradient Episodic Memory}
\label{subsec:gem}
The key challenge of continual learning is catastrophic forgetting, i.e. an agent forgets what it learned previously when adapting to a new environment. The phenomenon is especially prominent when feature-rich parametric models, i.e. neural networks, are used as the underlying learning module. To address catastrophic forgetting, we use Gradient Episodic Memory (GEM)~\cite{lopez2017gradient} within our Lifelong Learning for Navigation paradigm described in Section~\ref{subsec::problem}. 
From a high-level perspective, GEM prevents forgetting by ensuring each update will not increase the loss on previous tasks. Note that GEM allows new experience to improve performance on old tasks.
Specifically, assume the agent has already seen environments up to $\mathcal{E}_{k-1}$ and the learned policy is $\pi_{\theta_{k-1}}$. GEM assumes the agent keeps a small memory buffer $\mathcal{B} = \{\mathcal{M}_i\}_{i < k}$ that, for each previous environment $\mathcal{E}_i$, stores a few exemplary data points $\mathcal{M}_i$. GEM then optimizes the following objective:
\begin{equation}
    \min_{\theta} \ell(\pi_{\theta}, \mathcal{E}_k), ~\text{s.t.}~\ell(\pi_{\theta}, \mathcal{M}) \leq \ell(\pi_{\theta_{k-1}}, \mathcal{M}), ~\forall \mathcal{M} \in \mathcal{B},
\end{equation}
where $\ell(\pi, X)$ is the loss function that evaluates performance of $\pi$ on data $X$.\footnote{Here we abuse the notation so that $\ell(\pi, \mathcal{E})$ refers to the loss evaluated on data generated from environment $\mathcal{E}$.} For instance, in a regression task where we aim to predict the regression label $a$ from a given state $s$, then $\ell(\pi_{\theta}, X) = \mathbb{E}_{(s, a) \sim X} ||\pi_{\theta}(s) - a||_2$.
To efficiently solve the above optimization, GEM observes that the constraints are satisfied as long as 1) the new $\theta$ is initialized from $\theta_{k-1}$, and 2) at each optimization step, the loss on previous tasks does not increase. Assume the optimization steps are small, we can determine whether a new update increases the loss on a previous task by computing the inner product between the gradients on the current and previous tasks.
For example, the loss on a previous task will only increase if the inner product is negative. The optimization problem then becomes
\begin{equation}
    \vspace{2pt}
    \label{eq:gem}
    \min_{\theta} \ell(\pi_{\theta}, \mathcal{E}_k), ~\text{s.t.}~ \langle \pdv{\ell(\pi_{\theta}, \mathcal{E}_k)}{\theta},\pdv{\ell(\pi_{\theta}, \mathcal{M})}{\theta}\rangle \geq 0, ~\forall \mathcal{M} \in \mathcal{B}.
    \vspace{2pt}
\end{equation}
In practice, to solve Equation (\ref{eq:gem}), GEM uses stochastic gradient descent with a modified gradient. In particular, denote $g=\pdv{\ell(\pi_{\theta}, \mathcal{E}_k)}{\theta}$ and $g_i = \pdv{\ell(\pi_{\theta}, \mathcal{M}_i)}{\theta},~~\forall i<k$. Then, GEM finds the update direction $\tilde{g}$ by solving:
\begin{equation}
    \label{eq:primal}
    \tilde{g} = \argmin_{z} ||g - z||_2,~\text{s.t.}~\langle z,g_i\rangle\geq 0,~~\forall i<k.
\end{equation}
The above optimization is already in a nice quadratic form, but the decision variable $\tilde{g}$ has the same dimension as $\theta$, which can be millions for deep architectures. Fortunately, its dual problem is only associated with $k-1$ variables and can be efficiently solved by standard quadratic programming solvers. Formally, the dual problem of (\ref{eq:primal}) is
$$
    v^* = \arg\max_v~v^TG^TGv + g^TGv~~\text{s.t.}~~v \geq 0 \in \mathbb{R}^{k-1},
$$
where $G = -[g_1, g_2, \dots, g_{k-1}]$ is a matrix having $-g_i$ as its columns. As a result, the final $\tilde{g} = G^Tv^* + g$ and the update rule is $\theta \leftarrow \theta - \alpha \tilde{g}$, where $\alpha$ is the learning rate. GEM has no requirement for dynamically expanding the parameter size and often requires very few exemplar data points from past experience to maintain the learned behavior. Therefore, GEM is particularly suitable for robot navigation tasks since mobile robots often have very limited onboard memory resources.
\section{Lifelong Learning for Navigation}
% \xuesu{We can probably mention two main points here to give the readers more high level guidance: (1) We don't learn the full navigation policy (we have $\pi_0$), but only where the robot has trouble. So we need to identify trouble (Discriminator) and how to overcome trouble ($A_{correct}$). This could be one paragraph. (2) As an online learning agent with computation and memory constraints, we can not memorize and learn from everything (three buffers). This could be another paragraph. This way, maybe we can remove the billeted list? }
Intuitively, catastrophic forgetting is caused by the dilemma of overwriting old knowledge when learning new things. In real-world navigation, it is unlikely that the agent needs to keep learning new things every second. The agent can easily navigate in many places with classical approaches and only have trouble navigating in particular scenarios (e.g. the green and red path segments in Figure \ref{fig::three_tasks}). As a result, we regard Lifelong Learning for Navigation (LLfN) as a framework that learns an auxiliary planner $\pi_\theta$ to assist a classical planner $\pi_0$ only for navigating those difficult instances. Moreover, updating $\pi_\theta$ should minimally influence the past learned behaviors. To achieve this goal, the agent should be able to first identify those ``difficult" scenarios, i.e. identify in what particular states $s$ does the agent generate suboptimal motion commands.
\begin{wrapfigure}{r}{0.53\columnwidth}
    \centering
    \includegraphics[width=0.53\columnwidth]{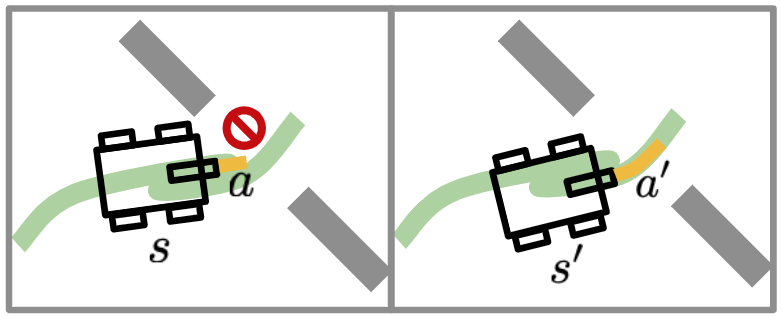}
    \caption{Learning from $(s', a')$ helps improve the navigation performance around $s$.}
    \label{fig:correct}
\end{wrapfigure}
In addition, since we do not assume access to expert demonstrations, the agent must keep sampling motion commands until it overcomes the difficulty around $s$. Then, by looking at the trajectories around $s$, the agent should be able to identify a good behavior, i.e. a state-action pair $(s', a')$, such that learning from $(s', a')$ will increase the agent's probability of overcoming the difficulty around $s$ (see Figure~\ref{fig:correct}).
% we focus the agent on learning from those difficult instances to reduce computation overhead and minimally influence its past knowledge. We would like the agent to be able to identify those ``difficult" scenarios from its experience, i.e. identify in what particular states $s$ does the agent generate sub-optimal motion commands. In addition, since we do not assume access to an expert to provide the ground truth motion commands for those difficult scenarios, the agent must keep sampling motion commands until it finds alternative better actions. 
% When it eventually passes a difficult scenario, it should identify the cause of its success, i.e. a particular motion command $a'$ that helped it navigate through the difficulty around $s$.
% Note that the motion command $a'$ is still generated by the agent itself, but might originate from a different state $s'$ near $s$ (see Figure~\ref{fig:correct}). The core assumption of lifelong learning from navigation (LLfN) is, by identifying the right $(s', a')$ and learning from it, the agent can improve its navigational behavior around $s$. Finally, the learning from $(s', a')$ should minimally affect what the agent has learned in the past.
Next, we list the key components of the LLfN framework and then explain how we use them.
\begin{itemize}
    \item An initial sampling-based navigation planner $\pi_0$ and a learnable policy $\pi_\theta$, parameterized by $\theta$.
    \item A scoring function $D: \mathcal{S}\times\mathcal{A} \rightarrow \mathbb{R}$ that evaluates how good an action $a$ is at the state $s$, i.e. larger $D(s, a)$ indicates $a$ is a better action at $s$. 
    \item A streaming memory buffer $\mathcal{B}_\text{stream}$ that stores the past $T$-step trajectory, i.e. $\mathcal{B}_\text{stream} = \{s_j, a_j\}_{j=t-T+1}^t$.
    \item A per-environment memory $\mathcal{M}_k: |\mathcal{M}_k| = n/k$ ($n$ is the memory budget) that stores the exemplar training data (self-generated data that are worth learning from) from environment $\mathcal{E}_k$. The entire memory before entering $\mathcal{E}_k$ is therefore a set of sets: $\mathcal{B} = \{M_i\}_{i<k}$.
    \item An algorithm $A_\text{correct}$ that given a recent suboptimal behavior $(s, a) \in \mathcal{B}_\text{stream}$, finds exemplar training data $(s', a') \in \mathcal{B}_\text{stream}$ such that learning from $(s', a')$ improves the navigation performance at $s$.
    \item A continual learning algorithm $A_\text{cl}$ that updates $\pi_\theta$ given $\mathcal{M}_k$ and $\mathcal{B}$. $A_\text{cl}$ should retain performance on previous environments.
\end{itemize}

With the above components, the agent will be manually placed at fixed start locations of a sequence of $m$ different environments, in each of which it navigates to a fixed goal, keeps identifying suboptimal behaviors, and improves upon them while preserving its past knowledge.\footnote{It may be possible for the agent to automatically detect environment shift, but we leave that for future work.} The pipeline of Lifelong Learning for Navigation is summarized in Algorithm~\ref{alg:clnav}.

\begin{algorithm}
    \caption{Lifelong Learning for Navigation (LLfN)}
    \label{alg:clnav}
    \begin{algorithmic}[1]
    \State \textbf{Inputs}: $\pi_0$, $\pi_\theta$, $D$, $A_\text{correct}$, $A_\text{cl}$, $\mathcal{E}_{k=1}^m$, and a threshold $\eta$.
    \State $\mathcal{B} \leftarrow \emptyset$, $\mathcal{B}_\text{stream} \leftarrow \emptyset$, and initialize $\theta_0$ randomly
    \State // Training
    \For{environment $k = 1 : m$}
        \State $\mathcal{M}_\text{k} \leftarrow \emptyset$
        \While{navigating in $\mathcal{E}_k$}
            \State progress to state $s_t$ and generate $a_t \sim \pi_0(s_t)$
            \State execute $a_t$ and update $\mathcal{B}_\text{stream}$ with $(s_t, a_t)$
            \State let $p = \floor*{t-T/2}$ and select $(s_p, a_p) \in \mathcal{B}_\text{stream}$
            \If{$D(s_p, a_p) < \eta$}
                \State $(s', a') = A_\text{correct}(s_p, \mathcal{B}_\text{stream})$
                \State update $\mathcal{M}_\text{k}$ with $(s', a')$
            \EndIf
        \EndWhile
        \State $\theta_k \leftarrow A_\text{cl}(\pi_{\theta_{k-1}}, \mathcal{M}_\text{k}, \mathcal{B})$ \algorithmiccomment{Lifelong learning}
        \State Shrink $\mathcal{B}$ to size $(n -|\mathcal{M}_\text{k}|)$ and $\mathcal{B}=\mathcal{B}\cup \{\mathcal{M}_k\}$
        % update $\mathcal{B}$ with $\mathcal{M}_k$
    \EndFor
    \State // Execution
    \While{navigating in $\mathcal{E}$}
            \State progress to state $s_t$
            \State generate $a_0 \sim \pi_0(s_t)$, $\hat{a} \sim \pi_{\theta_k}(s_t)$
            \State execute $a_t = \argmax_{a \in\{a_0, \hat{a}\}} D(s_t, a)$
    \EndWhile
    \end{algorithmic}
\end{algorithm}
In Algorithm \ref{alg:clnav}, the history memory buffer $\mathcal{B}$ and the streaming memory buffer $\mathcal{B}_\text{stream}$ are initialized to empty, and $\pi_{\theta}$ to a random policy in line 2. Upon entering $\mathcal{E}_k$ (line 4), the per-environment memory $\mathcal{M}_\text{k}$ is initialized to empty (line 5). While learning to navigate in $\mathcal{E}_k$ (line 6), when the agent progresses to a state $s_t$ (line 7), it executes the motion command $a_t \sim \pi_0(s_t)$ and saves the state-action pair $(s_t, a_t)$ to the streaming buffer $\mathcal{B}_\text{stream}$ (lines 7-8). Then, the agent looks at the mid-point\footnote{By looking at the midpoint $(s_p, a_p)$, we search the neighboring behaviors (within $T/2$ steps) for $(s', a')$.} of the recent trajectory $(s_p, a_p) \in \mathcal{B}_\text{stream}$ (line 9) and evaluates its score $D(s_p, a_p)$ (line 10). If the score is below a pre-defined threshold $\eta$, then we regard $(s_p, a_p)$ as suboptimal and use $A_\text{correct}$ to identify a nearby state-action pair $(s', a') \in \mathcal{B}_\text{stream}$, learning from which could potentially help improve the navigation performance around $s$ (line 11). In practice, we implement $D$ with the extracted heuristics from DWA which discriminates recovery behaviors from regular ones (see Section \ref{subsec::implementation} for details). In other words, suboptimal actions are the ones that trigger DWA's recovery behavior.
% In practice, we use the recovery behavior heuristics extracted from the DWA planner \cite{zheng2017ros} as $D$ as it naturally provides a good evaluation of navigational behavior (see Section \ref{subsec::implementation} for details).
% \xuesu{In practice, we implement a $D$, which discriminates the recovery behaviors produced (turn in place or back up) by DWA as bad actions whereas regular behaviors as good ones. }

Intuitively, if we search state-action pairs around $(s_p, a_p)$ that have scores above the threshold $\eta$, and find the state-action pair $(s', a')$ that has the most similar state to $s_p$, then learning from $(s', a')$ could potentially help the agent navigate from $s_p$. Therefore, we propose that $A_\text{correct}$ selects $(s', a')$ according to
\begin{equation}
    (s', a') = \argmax_{(s,a)\sim\mathcal{B}_\text{stream}} \text{sim}(s, s_p)~~\text{s.t.}~~D(s, a) \geq \eta,
\end{equation}
where $\text{sim}(s, s_p)$ measures how similar $s$ is to $s_p$. What we really want is to learn from $(s, a^*)$, where $a^*$ is the optimal motion command at state $s$ and is unknown by the policy. Here, $(s', a')$ serves as the nearest neighbor to $(s, a^*)$. The underlying intuition is that taking the same action in similar states should result in similar scores. Inversely, the similarity score indicates how confident we are that learning from the nearest neighbor $(s', a')$ can actually help the agent successfully navigate $s_p$, provided that $D(s', a') \geq \eta$. In our implementation, the state consists of raw sensor readings, such as LiDAR, so the negative Euclidean distance $-||s' - s_p||_2$ is a reasonable measure of similarity.
Importantly, although it is possible that within $\mathcal{B}_\text{stream}$, no state-action pair has sufficiently similar state to $s_p$, learning from the nearest neighbor $(s', a')$ is not detrimental to $\pi_\theta$ since $D(s', a') \geq \eta$.
% \xuesu{from Peter: This terminology of “learn from (s’,a’)” is a bit confusing.  What you mean exactly is “storing (s’,a’) in the ***” where *** is some term that we should use.  It’s not a training example for a generalization algorithm, but rather an instance used for nearest neighbor learning.  I guess in that sense we’re learning from it.  But we should use the “nearest neighbor” terminology.  I forget what the stored instances are called in that case, but we should use the same terminology.}
In line 12, $(s', a')$ is stored to $\mathcal{M}_k$. Note that $\mathcal{M}_k$ is also size-constrained, i.e. $|\mathcal{M}_k| \leq n/k$. So if adding $(s', a')$ exceeds the memory budget, we remove the $(s', a') \in \mathcal{M}_k$ with the \emph{lowest similarity score} to their own corresponding $s$. By doing so, $\mathcal{M}_k$ eventually consists of state-action pairs that most likely contribute to improvement of navigation performance.
% \xuesu{from Peter: Lowest similarity to what?  These points are supposed to be useful for navigation from everywhere in the environments, so we shouldn’t just be considering similarity to the current point, right?}
%
In line 15, the agent updates its knowledge with the current $\mathcal{M}_\text{k}$, using the continual learning algorithm $A_\text{cl}$, while preserving what it has learned before by considering $\mathcal{B}$. In particular, to continually update $\pi_\theta$, we adopt the Gradient Episodic Memory (GEM) as $A_\text{cl}$. Specifically, we would like to solve
\begin{equation}
    \label{eq:clnav}
    \begin{gathered}
    \theta_k = \argmin_\theta \ell(\pi_\theta, \mathcal{M}_\text{k}) ~~\text{subject to} \\
    \langle \pdv{\ell(\pi_{\theta}, \mathcal{M}_\text{k})}{\theta},\pdv{\ell(\pi_{\theta}, \mathcal{M})}{\theta}\rangle > 0,~~\forall \mathcal{M} \in \mathcal{B}.
    \end{gathered}
    \vspace{5pt}
\end{equation}
Here $\ell(\pi_\theta, \mathcal{M}) = \mathbb{E}_{(s,a) \in \mathcal{M}} ||a - \pi_\theta(s)||_2$, is the standard behavior cloning objective. The optimization in Equation~(\ref{eq:clnav}) will then be solved with stochastic gradient descent using the same method shown in Equation~(\ref{eq:primal}). Finally, to accommodate the online memory budget, we shrink $\mathcal{B}$ by removing entries with the lowest similarity score to their own corresponding $s$ and append data from $\mathcal{M}_k$ to $\mathcal{B}$ (line 16).

For executing the learned policy from LLfN, the only change is that given the state $s_t$, we compute both $a_0 \sim \pi_0(s_t)$ and $\hat{a}\sim \pi_\theta(s_t)$ and execute the action with the highest score (lines 19-23). For description simplicity, we learn once per environment in Algorithm~\ref{alg:clnav}. However, in practice, the method can be easily extended to learn multiple times while navigating in the same environment.
\section{Experiments}
\label{sec::experiments}
LLfN is tested in simulated and physical experiments. We hypothesize that through LLfN (1) navigation performance can improve as the robot gathers more experience within a single environment, and (2) navigation in new environments can be learned while not forgetting how to navigate well in previous ones. 

\subsection{Robot Platform and Implementation}
\label{subsec::implementation}
Clearpath Jackal, a four-wheeled differential-drive unmanned ground vehicle, is used for both simulated and physical experiments. The robot is equipped with a laser scanner to perceive surrounding obstacles and runs the basic Robot Operating System (ROS) \texttt{move\textunderscore base} navigation stack. While the global planner is based on Dijkstra's algorithm \cite{dijkstra1959note}, the local planner uses DWA \cite{fox1997dynamic}, a sampling-based planner which may fail to sample feasible actions if not being properly tuned for the deployment environment. We fix the global planner and improve the local DWA planner using LLfN. We include the local goal provided by the global planner as part of the state, along with LiDAR input. Based on the local DWA planner as an initial policy ($\pi_0$), LLfN learns to complement this $\pi_0$ and improve navigation performance in multiple environments. When the sampling-based DWA cannot find feasible motion, it starts recovery behavior, including rotation in place or backing up \cite{zheng2017ros}. LLfN first improves navigation by eliminating those sub-optimal behaviors and producing alternative motions for a given environment. Second, it allows the robot to adapt to new environments while still remembering previous ones. 
% \peter{Have we introduced the notion of a global and local planner yet in this paper?  Does global appear in the algorithm?}

The learning problem is formulated as finding a policy $\pi_\theta$ that maps from the current state $s$, which includes LiDAR input (720- and 2095-dimensional for simulated and physical experiments, respectively) and local goal ($\left(x, y\right)$, 1m away on the global path), 
% \commentp{that's too informal a specification of the state space.  Be precise, give the dimensionality, etc.} 
to the action $a$, i.e. the linear velocity $v$ and angular velocity $\omega$.
In our implementation, the scoring function $D$ prioritizes actions proposed by DWA, if they result in a steady forward motion ($v\geq0.15\text{m/s}$, which is slightly larger than the minimal $v=0.1\text{m/s}$ to overcome friction). Importantly, all recovery behaviors from the DWA planner, including slowing down, turning in place, and moving backward, result in $v < 0.15\text{m/s}$. So with the above definition, $D$ will classify all recovery behaviors as suboptimal.
% This choice of implementation is due to the fact that DWA with default parameters is a relatively conservative algorithm, which only produces steady forward motion when it thinks the motion is absolutely safe.
Future works can investigate other implementations of $D$, or even learning it on the fly. If DWA fails to find feasible actions and would have started to execute recovery behaviors, the learned policy $\pi_\theta$ takes over. We acknowledge this implementation makes the scoring function $D$ not always accurate, e.g. it is possible that $D$ may prefer a bad action by $\pi_\theta$ when $\pi_0$ produces recovery behaviors. But in practice we observe LLfN remains effective even with such a simple $D$.

In consideration of limited onboard resources, we implement a streaming buffer $\mathcal{B}_\text{stream}$ as a regular queue to store 300 online streaming data points sequentially, and a separate memory $\mathcal{M}_\text{k}$ as a priority queue to save exemplar training data in the current environment. The number of data points within $\mathcal{B}\cup \{\mathcal{M}_k\}$ is also constrained to 300. The buffer sizes are empirically determined as the minimal value to assure efficient learning. For every sub-optimal action generated by DWA at a certain state, we compute a similarity score between this state and all states in the streaming buffer with a successful action (L2-norm of the LiDAR reading difference). We replace the data point of lowest similarity score in the training buffer with that from the streaming buffer.
% For duplicated entries, we simply update the similarity score.
% \peter{This detail also seems like it should have been introduced earlier.  The actually similarity score is an implementation detail.  But the way you get training data and trade off between the controller and the learned policy are aspects of the general algorithm, and should be introduced earlier.} \xuesu{see the comment above to add one more component}
The onboard data memory overhead to implement LLfN is \emph{at most} a total of 600 data points. 

\subsection{Simulated Experiments}
\begin{figure}
\centering
\subfloat[Env. 1]{\includegraphics[height=1.33cm]{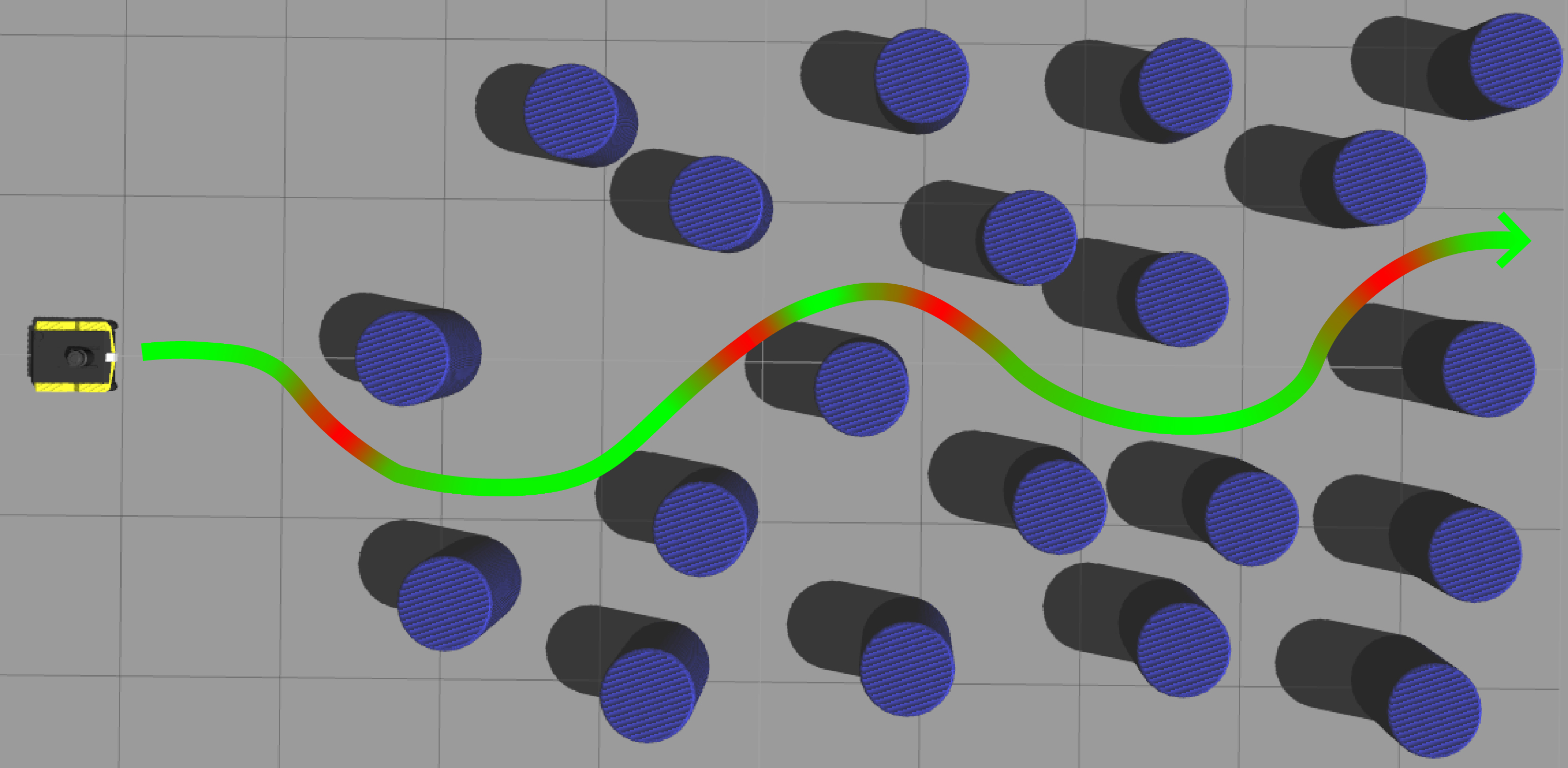}%
\label{fig::task1}}
\hspace{2pt}
\subfloat[Env. 2]{\includegraphics[height=1.33cm]{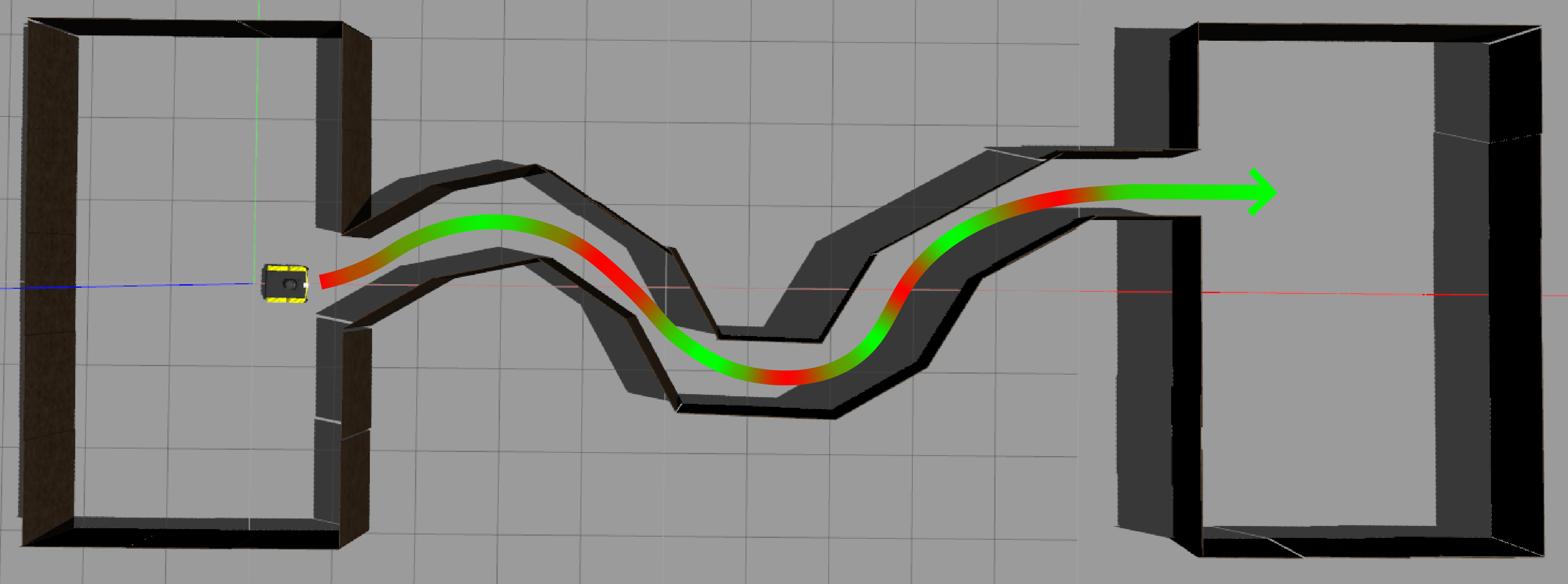}%
\label{fig::task2}}
\hspace{2pt}
\subfloat[Env. 3]{\includegraphics[height=1.33cm]{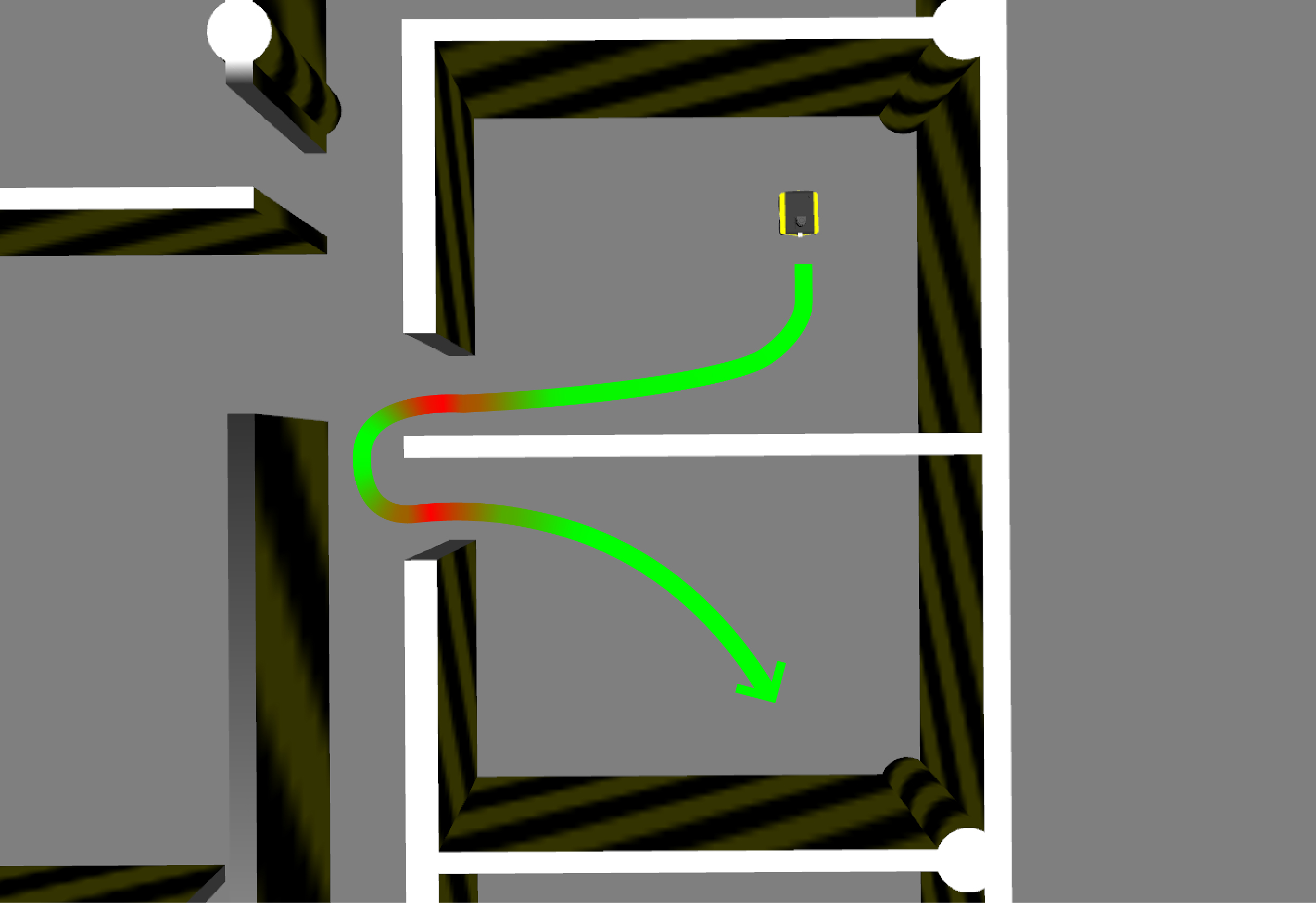}%
\label{fig::task3}}
\caption{Simulated Navigation Environments: Green segments are primarily traversed using the initial policy $\pi_0$, while red segments are mostly traversed using the learned planner $\pi_\theta$. }
\label{fig::sim_tasks}
\vspace{-10pt}
\end{figure}

The effect of LLfN is first studied with extensive simulated trials. The simulated Jackal has a SICK LMS111 laser scanner onboard providing 270$^{\circ}$ 720-dimensional laser scan. The three simulated navigation environments are shown in Figure \ref{fig::sim_tasks}, where the robot navigates from a fixed start to a fixed goal in the environment. 
Env. 1's dense obstacles require fast response for obstacle avoidance; In Env. 2, the robot needs to keep a slow and consistent pace to drive through the narrow passage; Env. 3 requires the robot to slow down to make a sharp turn to enter the other room smoothly.

The robot is placed at its start location in each environment, and the learned policy $\pi_\theta$ to be tested is selected manually. 
% \commentp{In the figure, it looks like the robot gets teleported from one environment to the next.  That's not the image I had from previous text and Figure 1 where it looks like there's some smooth transition from one environment to the next, but somehow the agent knows when it happens.  Also, remember you said the agent goes from one "end" of an environment to another, but that's not what it looks like in this figure.}

We use the initial policy $\pi_0$ to execute three trials in each of the three navigation environments to collect the training data. $\pi_0$ (DWA) can eventually navigate through, given sufficient time to recover, re-sample, and re-plan. 
% \commentp{This assumes that the initial policy can succeed at navigating through the whole environment, right?} 
To test in-environment learning improvement, training data of each given environment is \emph{divided into} 5 segments and incrementally presented to the learner. 
For example, the first training buffer for learning is constructed from the first one-fifth of the robot's experience, the second from the first two-fifths, etc. The last training buffer is from the entire experience. Note all these training buffers contain only 300 data points, who have the highest similarities to those states where the initial policy performs sub-optimally, given the presented navigation experience. 
% This results in training buffers of 300\peter{I don’t understand what you’re saying here.} data points with the highest similarities to those observations where the initial policy performs sub-optimally, given the presented navigation experience. 
Therefore, 6 incrementally learned navigation policies are produced, including the initial policy and the five policies learned from seeing 1/5, 2/5, ..., 5/5 of the training data from three training trials. Then the robot starts to learn the next navigation environment. For cross-environment learning, we use Sequential Training and the proposed LLfN. Sequential Training first uses the training buffer of environment 1. Starting with the final policy trained from environment 1, it sequentially trains on the training buffer of environment 2, before moving on to the training buffer of environment 3. In LLfN, while training environment 2, it only uses 150 data points from training buffer 2 with the highest similarity score, and still keeps a memory of 150 data points with the highest similarity score from training buffer 1 to assure new gradient updates won't increase the loss of environment 1. While training on environment 3, 100 data points with the highest similarity score for each environment are used, to avoid forgetting environment 1 and 2. 
\begin{figure}
\centering
\includegraphics[width=\columnwidth]{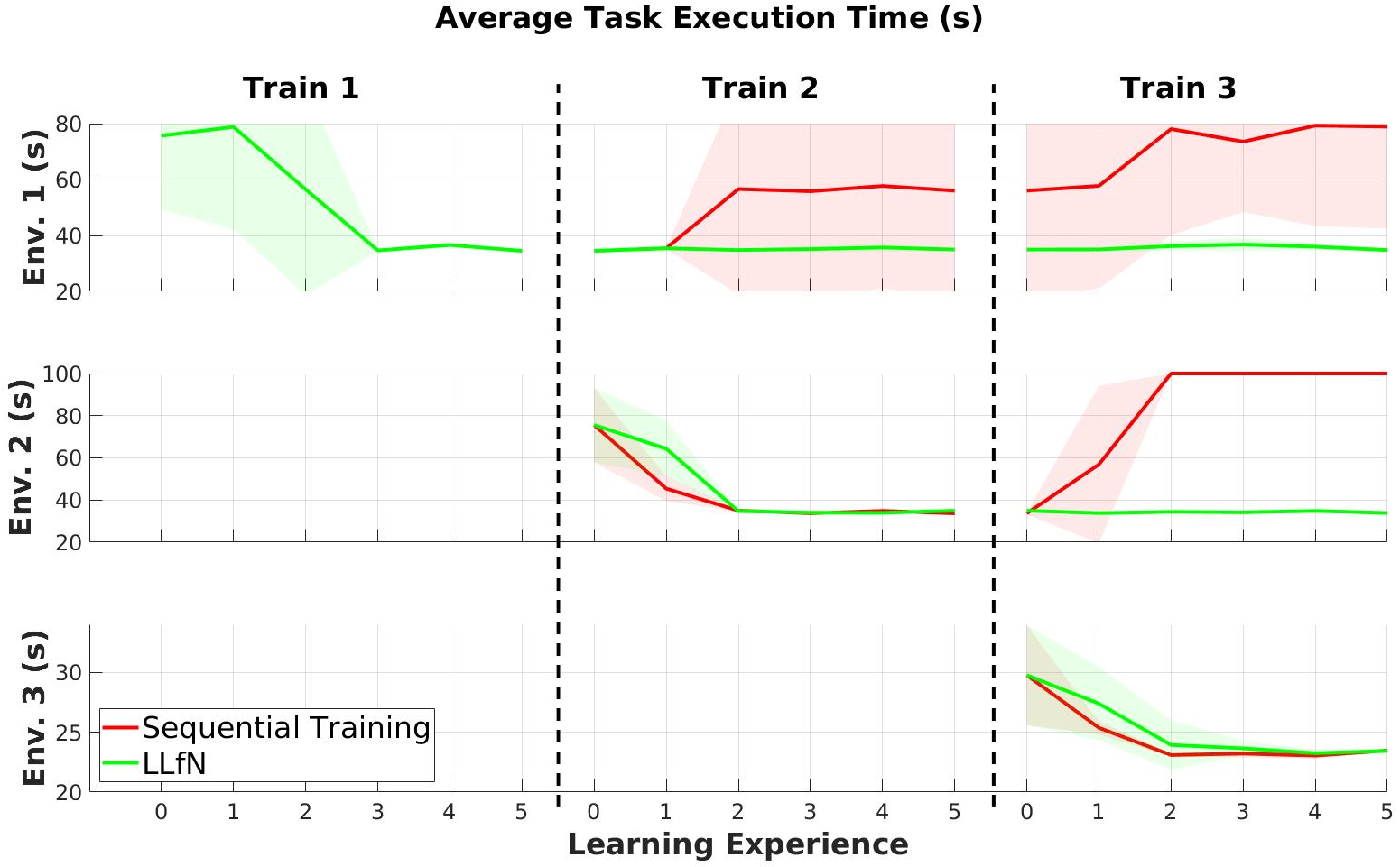}
\caption{Simulation Results: Each subplot (row $i$, column $j$) shows navigation performance with respect to increasing experience (x-axis), being trained using the $j$th and deployed in the $i$th environment. (Since there is no previous environment for Env. 1 and Train 1, both curves are the same.)}
\label{fig::sim_test}
\vspace{-10pt}
\end{figure}

We implement a small neural network of three hidden layers with 64 hidden neurons each, to compute linear and angular velocity based on the LiDAR input and local goal. After training, we evaluate the navigation performance in terms of traversal time in the current and previous environments. Each policy is executed three times, resulting in a total of 198 evaluation trials. We report the mean and standard deviation of the performance in Figure \ref{fig::sim_test}. If the robot is not able to reach the goal, e.g. gets stuck, a penalty time of 100s is given. Within each environment, LLfH is able to decrease traversal time with increasing learning experience. Across environments, LLfH learns new environments with increasing data while avoids catastrophic forgetting of previous ones. Sequential Training can improve navigation performance of a given environment when being presented the data from that particular environment, faster than LLfN. However, navigation performance in previous environments deteriorates with increasing experience in the current environment. The catastrophic forgetting is apparent in the diverging red line (Sequential Training) from the green line (LLfH). Note that the learned policies are used in conjunction with the initial policy in both cases. 

\subsection{Physical Experiments}
LLfN is also implemented on a physical Jackal robot. In the physical experiments (Figure \ref{fig::three_tasks}), we use the same setup, while all computation is done onboard the robot using an Intel Core i5-4570TE CPU. The physical Jackal has a Velodyne LiDAR, whose 3D point cloud data is converted to 360$^{\circ}$ 2095-dimensional laser scan. We design the physical test environments such that the goal of one environment smoothly transitions to the start of the next. Therefore the robot can traverse through all three environments in one shot. For training, we manually label the current environment for the learner. During deployment, the robot uses only one $\pi_\theta$ for each traversal. 

In the physical experiments, we compare LLfN with three baselines: DWA, Sequential Training, and Individual Models.
% \peter{That’s the right way to say it.  But then in Table 1, it again looks like GEM is what’s being compared.} 
Sequential Training and LLfN are conducted in the same way as in the simulated experiments. Individual Models are checkpoints of Sequentially Trained models up to the corresponding environment. Since these Individual Models do not suffer from forgetting, they serve as the \emph{best} models in Sequential Training. In contrast, the Sequentially Trained model is the final model after training on all 3 environments (thus ends up being much more effective on Environment 3 than Environment 1 or 2). The downside of Individual Models is that the number of models increases with the number of environments. Thanks to the constraint on data size, training each model takes less than two minutes on the robot's onboard CPU.
% \peter{The other thing that comes across to me gradually - really only in the experimental results section - is the fact that the default controller is used most of the time and the learned model is very sparsely used.  Presumably that’s why you can get away with so little data.  It’s a very specialized model.  I still find it confusing how much data is used to train the discriminator and the control policy, and exactly how they interact.  Actually, I’m now not sure that you have 2 different models.  I will go back and try figure it out.  But it needs to come out more clearly.  Another thing that could be useful to show is where along the trajectories in the environments the default controller and the learned policy are used.  How often is the policy invoked?  That would have to go in supplementary material, but at least in simulation, is it possible to illustrate?}
% \begin{wraptable}{r}{0.6\textwidth}
\begin{table}[t]
\centering
\scalebox{0.85}{
\begin{tabular}{cccccc}
\toprule
& & \textbf{DWA} & \textbf{\begin{tabular}[c]{@{}c@{}}Sequential \\ Training\end{tabular}} & \textbf{\begin{tabular}[c]{@{}c@{}}LLfN\end{tabular}} & \textbf{\begin{tabular}[c]{@{}c@{}}Individual\\ Models\end{tabular}} \\ \midrule
\multicolumn{1}{c}{\multirow{2}{*}{\textbf{Env. 1}}} & Time & 38.36$\pm$1.69  & 39.11$\pm$6.45 & \textbf{28.32}$\pm$1.85  & 30.17$\pm$\textbf{0.97}\\ 
\multicolumn{1}{c}{} & Rec./Col. & 1.8/0 & 1.4/0.6 & 0/0 & 0/0\\ \midrule
\multicolumn{1}{c}{\multirow{2}{*}{\textbf{Env. 2}}} & Time               & 28.05$\pm$3.48 & 49.41$\pm$ 13.94 & \textbf{19.98}$\pm$1.07 & 23.41$\pm$\textbf{0.66}\\
\multicolumn{1}{c}{} & Rec./Col. & 1.4/0.2 & 1.25/0.8 & 0/0 & 0/0 \\ \midrule
\multicolumn{1}{c}{\multirow{2}{*}{\textbf{Env. 3}}} & Time & 39.80$\pm$6.39  & 21.12$\pm$1.17 & 21.80$\pm$1.51 & \textbf{21.12}$\pm$\textbf{1.17} \\ 
\multicolumn{1}{c}{} & Rec./Col. & 2.2/0.2 & 0/0 & 0/0 & 0/0\\ \bottomrule
\end{tabular}}
\caption{~Physical Results after Training on 3 Environments: Time (in s), Number of Recovery Behaviors (Rec.) and Collisions (Col.).}
\vspace{-5pt}
\label{tab::experiments}
\end{table}
We deploy the trained models to navigate in the three environments. For each method, the robot navigates each environment five times, resulting in a total of 60 physical trials. Table \ref{tab::experiments} reports the mean execution time for each environment with standard deviation and average number of recovery behaviors (Rec.) and collisions (Col.). 
DWA exhibits the most recovery behaviors, because whenever the robot fails to sample a feasible motion, it starts recovery behavior. One collision happens in one of the environment 2 and 3 trials. Applying the model sequentially learned on environment 1, 2, and 3 causes catastrophic forgetting of the first two environments. It leads to longer execution time and higher standard deviation, with frequent recovery behaviors and collisions. LLfN can successfully avoid catastrophic forgetting: It achieves a similar time to the Individual Models approach in environment 3, while, \emph{surprisingly}, it outperforms Individual Models in environment 1 and 2 in terms of average time. One possible explanation is that the LLfN model has good backward transfer ability after gathering more diverse experience.
Utilizing extra training data and models specifically trained for each environment, the Individual Models approach is more stable (lowest standard deviation). (Video of representative trials of the four methods: \url{www.youtube.com/watch?v=ja_Rjc63xiY&t=68s}.)

\section{Conclusion}
\label{sec::conclusions}
In this paper, we propose and implement the first self-supervised Lifelong Learning for Navigation framework (LLfN). Building upon an initial static sampling-based model predictive control policy, which does not improve with increasing navigation experience, the robot is able to self-identify sub-optimal actions, search for similar scenarios where good actions are performed, learn from those data, and improve navigation in a continual manner. Furthermore, in a multi-environment setting, LLfN is able to adapt to new environments, while not forgetting previous ones. Extensive simulated trials are performed to test LLfN's in-environment and cross-environment learning capability. The entire LLfN is also implemented and tested on limited computational resources onboard a physical robot and operates in real time without requiring any off board computation.
One interesting future direction is to extend the current simple heuristic-based scoring function to a more general formulation, e.g. in terms of a learnable value function. Then the scoring function will not depend on heuristics specific to the initial policy (DWA in our case). One shortcoming of the current framework is that when the base policy $\pi_0$ is incompetent everywhere, LLfN will fail to learn. Therefore future research can investigate adding active exploration when $\pi_0$ is always incompetent. Other interesting directions include investigating better methods to prioritize experiences for updating the online buffers with a limited budget and extending LLfN to address dynamic obstacles. 
% \peter{Add a sentence or two of possible future work. }

\section*{ACKNOWLEDGMENTS}
This work has taken place in the Learning Agents Research Group (LARG) at UT Austin.  LARG research is supported in part by NSF (CPS-1739964, IIS-1724157, NRI-1925082), ONR (N00014-18-2243), FLI (RFP2-000), ARO (W911NF-19-2-0333), DARPA, Lockheed Martin, GM, and Bosch.  Peter Stone serves as the Executive Director of Sony AI America and receives financial compensation for this work.  The terms of this arrangement have been reviewed and approved by the University of Texas at Austin in accordance with its policy on objectivity in research.
%%%%%%%%%%%%%%%%%%%%%%%%%%%%%%%%%%%%%%%%%%%%%%%%%%%%%%%%%%%%%%%%%%%%%%%%%%%%%%%%

\bibliographystyle{IEEEtran}
\bibliography{IEEEabrv,references}

\end{document}